\documentclass[a4paper]{svmult}

\usepackage{type1cm}        
\usepackage{times}
\usepackage{algorithm}
\usepackage{multirow}
\usepackage{amsmath}
\usepackage{amsfonts}
\usepackage{algpseudocode}
\usepackage{color}
\usepackage{xcolor}
\usepackage{bm}
\usepackage{tabularx}
\usepackage{subcaption}
\usepackage{url}
\usepackage{cite}
\usepackage{prettyref}
\usepackage{booktabs}
\usepackage{siunitx}

\usepackage{makeidx}         
\usepackage{graphicx, epstopdf}        
\usepackage{multicol}        
\usepackage[bottom]{footmisc}

\usepackage[numbers]{natbib}

\usepackage{newtxtext}       %
\usepackage{newtxmath}       

\usepackage{hyperref}
\hypersetup{
    colorlinks,
    linkcolor={blue!80!black},
    citecolor={blue!80!black},
    urlcolor={blue!100!black}
}

\usepackage[compact]{titlesec} 

\newrefformat{fig}{Fig.~\ref{#1}}
\newrefformat{sec}{Sec.~\ref{#1}}
\newrefformat{tab}{Tab.~\ref{#1}}
\newrefformat{alg}{Algo.~\ref{#1}}
\newrefformat{eqn}{Equ.~\ref{#1}}

\makeindex             

\begin{document}

\title*{Crowd-Driven Mapping, Localization and Planning}

\author{Tingxiang Fan, Dawei Wang, Wenxi Liu and Jia Pan}

\institute{Coresponding author: Jia Pan \at the University of Hong Kong, HKSAR  \email{jpan@cs.hku.hk}. This work is partially supported by Innovation and Technology Fund (ITF) ITS/457/17FP, and General Research Fund (GRF) 11207818, 11202119, NSFC-RGC joint grant N\_HKU103/16.}

\maketitle

\pagestyle{empty} 

\vspace{-3.5cm}

\abstract{Navigation in dense crowds is a well-known open problem in robotics with many challenges in mapping, localization, and planning. Traditional solutions consider dense pedestrians as passive/active moving obstacles that are the cause of all troubles: they negatively affect the sensing of static scene landmarks and must be actively avoided for safety. In this paper, we provide a new perspective: the crowd flow locally observed can be treated as a sensory measurement about the surrounding scenario, encoding not only the scene's traversability but also its social navigation preference. We demonstrate that even using the crowd-flow measurement alone without any sensing about static obstacles, our method still accomplishes good results for mapping, localization, and social-aware planning in dense crowds. Videos of the experiments are available at \url{https://sites.google.com/view/crowdmapping}.}

\section{Introduction}

Navigating a mobile robot in complex, cluttered, and dynamic human scenarios has a wide variety of applications, including last mile delivery, service robot in malls, and surveillance. However, most existing navigation algorithms only work well in scenarios with only a few or no person due to two main challenges in dense crowds.

\begin{figure}
\centering
\begin{subfigure}{0.49\textwidth}
\centering
\includegraphics[width=1.0\linewidth]{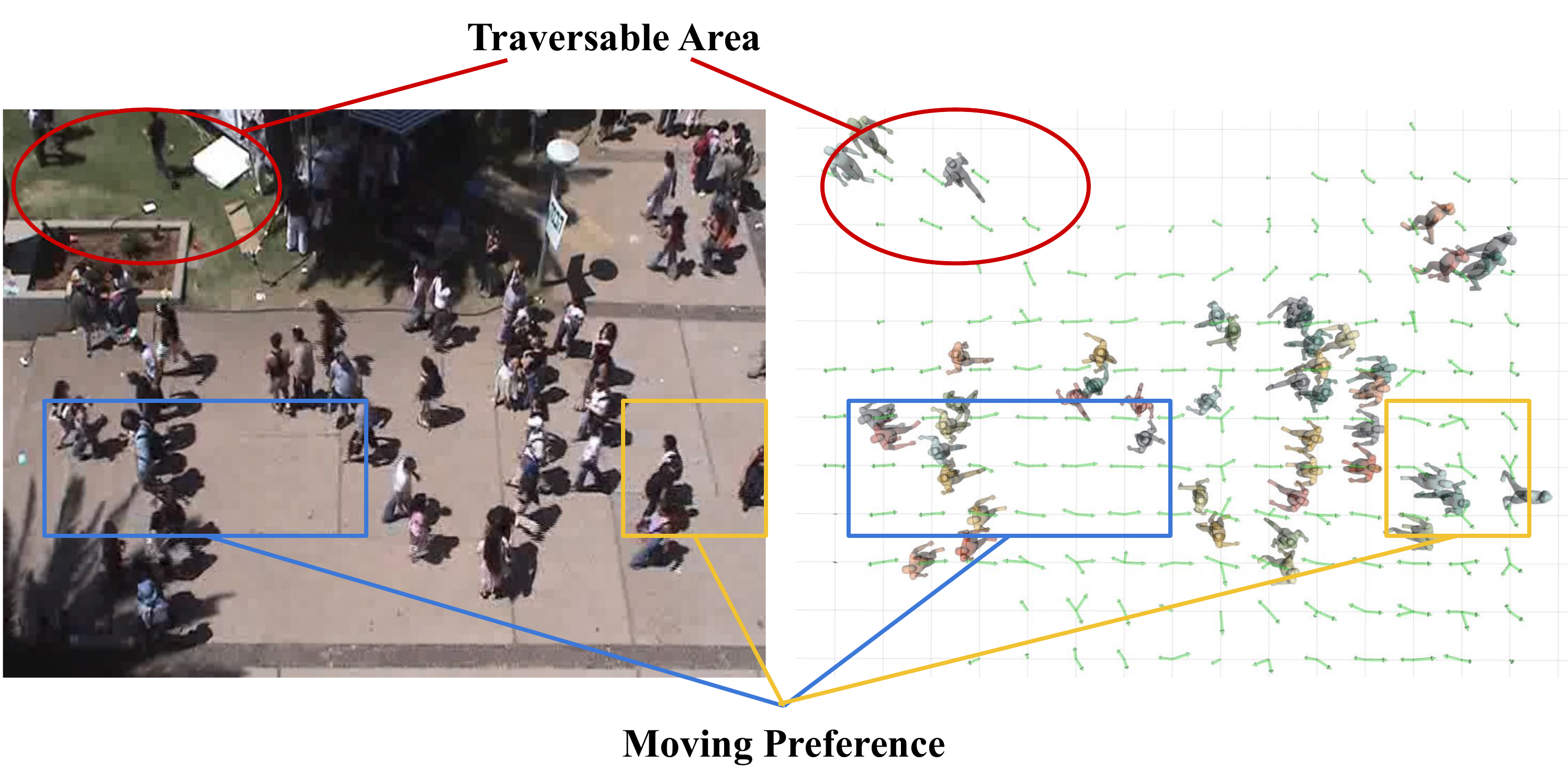}
\caption{a video clip from UCY dataset~\cite{UCY_Dataset}}
\end{subfigure} 
\begin{subfigure}{0.5\textwidth}
\vspace{0.4cm}
\centering
\includegraphics[width=0.93\linewidth]{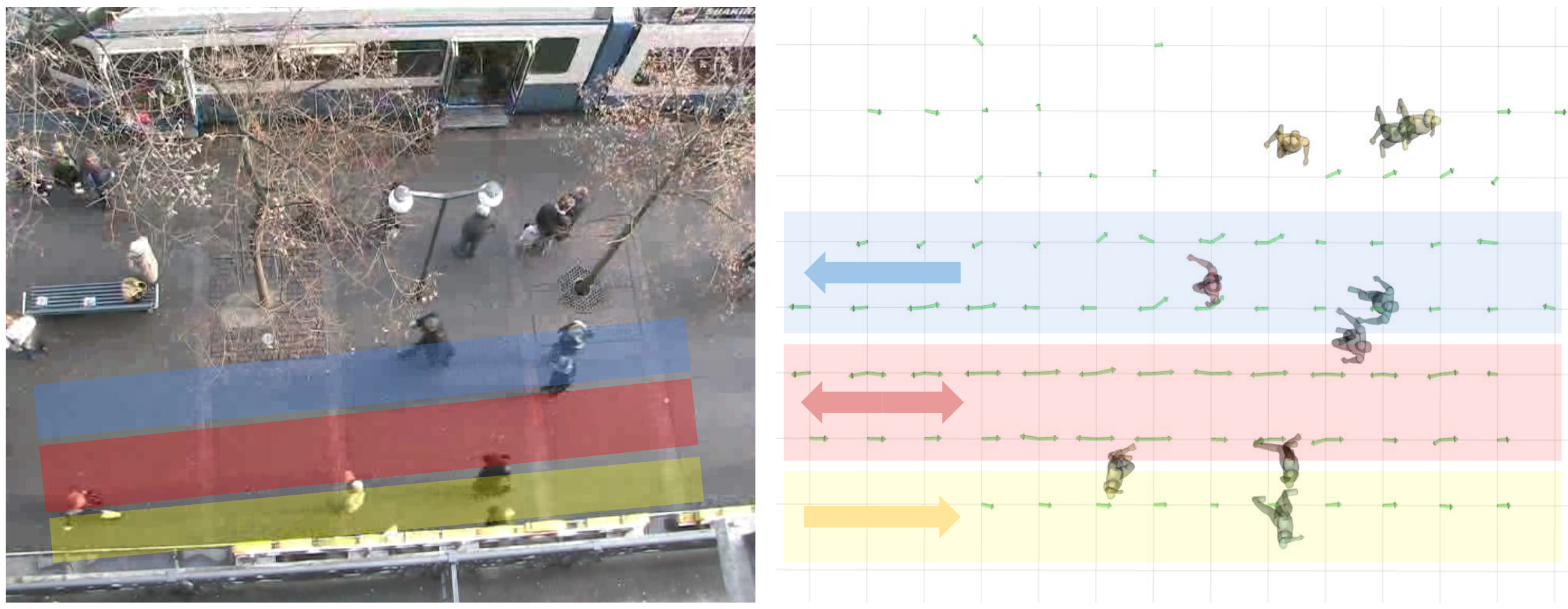}
\vspace{0.32cm}
\caption{a video clip from ETH dataset~\cite{ETH_Dataset}}
\label{fig:eth_hotel}
\end{subfigure}
\caption{Two real-world examples illustrate two types of latent social information in crowd flow: (a) crowd flow indicates the region traversability and scene's moving preference; (b) crowd flow encodes the scene's social navigation rules.}
\label{fig:cover}
\vspace{-5pt}
\end{figure}

First, in a dense crowd, a robot is difficult to figure out the layout of surrounding environment and its accurate location using the classical simultaneous localization and mapping (SLAM) framework, because the existence of moving obstacles like pedestrians can occlude the robot's observation and the landmark features may be confused with features on moving objects, resulting in mistakes in correspondence calculation. Thus, prior works often treat moving objects as interference and intend to filter them out from the sensory observation, which unfortunately is difficult to be implemented robustly even with the recent progress in computer vision. Such filtering also discards the rich human-robot interaction information that is helpful for achieving social-friendly robot behaviors.

Second, in a dense crowd, a robot is difficult to accomplish safe and effective navigation without getting stuck in crowds.
Some recent works consider pedestrians as moving obstacles and use either reactive policies~\cite{long2018towards, fan2019getting,everett2018motion} or short-horizon planning~\cite{Cai:2019:LETS} to trade-off collision avoidance and goal approaching. However, these methods do not model social-awareness and thus the resulting robot motion may be safe but intimidating, e.g., the robot may abruptly accelerate or stop towards its neighbors' tiny movements. Such overly responsive behavior will disturb the normal crowd behavior and significantly reduce the robot's social-friendliness and navigation effectiveness.

In this paper, we solve above challenges by repurposing the role of pedestrians in autonomous robotic navigation. Rather than treating pedestrians as noises to static landmarks or obstacles to be actively avoided, we use the observation of nearby pedestrian crowds as another type of sensory measurement about the surrounding environment with crowds. In particular, we consider the following navigation problem in the extreme case: suppose the robot has no sensor mechanism for measuring geometry of static obstacles and can only perceive the movement of its nearby pedestrians, can the robot accomplish mapping, localization, and planning? Motivated by human’s behaviors in \prettyref{fig:cover}, we find the answer is YES, because the observed vector field of local crowd flow can be used to infer the \textit{traversability} of different locations in the unstructured scene and the crowd's social \textit{moving preferences}, including rules of left/right hand traffic, crossing and overtaking~\cite{chen2017socially}. These two types of crowd latent information are of great importance for achieving mapping, localization, and social-friendly navigation in dense crowd scenes.

{\noindent \bf Main Contributions:} We propose a novel framework, called \textit{CrowdMapping}, for robotic navigation in dense crowds. It includes three modules: 
\begin{itemize}
\item \textit{CrowdMapper} uses local crowd observation as sensory measurement and applies clustering algorithm to reconstruct a crowd-flow map capturing the movement pattern of pedestrians in the scenario;
\item \textit{CrowdLocalizer} uses a flow-matching metric to leverage the crowd-flow map for localization;
\item \textit{CrowdPlanner} allows a robot to follow the movement pattern in the crowd-flow map and accomplish social-aware planning. 
\end{itemize}

\section{Related Work}

\subsection{Crowd-Flow Estimation}
To capture the crowd-flow pattern of the environment, many estimation methods have been developed. According to the observation perspective, we categorize existing crowd-flow estimation methods into two types, global and local methods. The global estimation methods use global cameras or drones to obtain observation of the crowd flow in the entire scene ~\cite{zhang2019wide, gajjar2017human}.  However, the global observation is often unattainable for robots due to the limited viewports of on-board sensors. 
Thus, local estimation methods are proposed to estimate the crowd flow around the robot.  Many exiting work have achieved real-time pedestrian flow tracking by on-board sensors~\cite{miller2016ped_tracking, linder2016multi_ped_tracking, yan2017online_ped_tracking}. Compared to the global estimation approaches, however, these local methods with limited perceptive are more likely to encounter the occlusion issue. To overcome it, some studies propose the more robust occlusion-aware estimation approaches for tracking systems~\cite{dong2016occlusion, judd2019occlusion_mvo}.

\subsection{Crowd-Flow Navigation}
Navigation in crowd-flow involves the complex human-robot interaction. Some previous works on crowd navigation usually consider pedestrians as dumb moving obstacles that need to be avoided actively~\cite{long2018towards,fan2019getting,chen2017decentralized}. Thus, the resulting robot behavior can be too sensitive to human's subtle movement, which may disturb the pedestrian flow and reduce the throughput of the entire crowd. 
To eliminate this issue, some methods formulate the social flow factors inside the human-robot interaction model. \cite{muller2008socially} presents an iterative A$^*$ algorithm to follow a pedestrian along the crowd flow, but it works only when the pedestrian to be followed and the robot share the same goal. In~\cite{henry2010learning}, the robot uses inverse reinforcement learning to imitate the pedestrian's behavior in a crowd for reducing the disturbance between the robot and the crowd. It uses Gaussian processes to online estimate the pedestrian flow vector field in the scenario, but is sensitive to the time-varying crowd mode and thus not appropriate for mapping and localization. More importantly, the mimic policy is not optimal in terms of minimizing the robot-crowd disturbance. \cite{aroor2019online} introduces a crowd-aware planner to plan a trajectory that conforms to the social-flow direction and also avoids congestion, but it assumes the global crowd flow field to be static and known beforehand, which unfortunately is often impractical.

\section{CrowdMapping Framework}

\begin{figure*}
\centering
\includegraphics[width=1.0\linewidth]{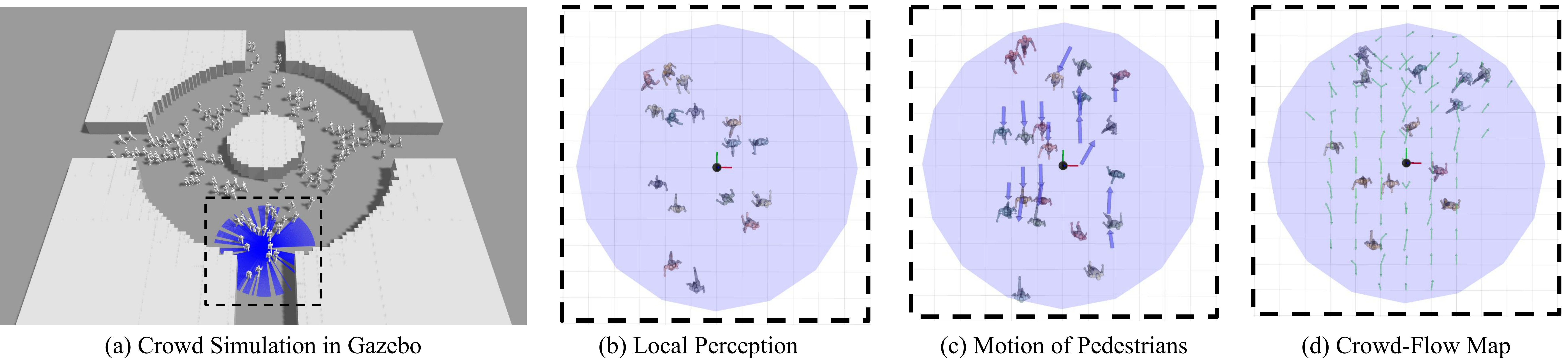}
\caption{The \textit{CrowdMapper} pipeline. (a): the simulated crowded scenario, with the local region around the robot highlighted; (b): the robot's observation, where the blue circle indicates the perception region; (c): the captured motion vector field of the robot's nearby pedestrians; (d): the recovered the time-invariant crowd-flow feature that will be fused into the global crowd-flow map. }
\label{fig:mapping_demo}
\end{figure*}

Our fundamental assumption is that the pedestrian crowd flow can provide sufficient time-invariant features for mapping the navigation scenario. In particular, the moving patterns in the crowd flow reflect the resistance of static obstacles to the crowd movement along different directions: the high resistance along a direction indicates the existence of obstacles along that direction and vice versa. Such resistance effect depends on map layout and pedestrians' moving preference, and thus is mostly time-invariant, making it suitable for map reconstruction. For instance, in the sidewalk scenario shown on the right of \prettyref{fig:cover}, there are two types of crowd flow patterns. In the yellow box region, there is no resistance in all directions, thus humans can move freely and their paths may intersect. In the blue box region, people prefer moving along the east-west direction, implying the existence of street boundary that resists human motion in the south-north direction. Such directional resistance can also be used to encode social navigation rules, e.g., the right-traffic rule in \prettyref{fig:eth_hotel}. Such crowd flow patterns are time-invariant and can be used to reconstruct a map reflecting the navigation constraints posed by the scenario obstacles and social rules.

We now describe our novel mapping framework and its applications in three steps. First, \prettyref{sec:crowd_mapper} presents the building of global crowd-flow map from the robot's local perception of the pedestrian crowd. The obtained crowd-flow map is then used for localization in  \prettyref{sec:crowd_localizer}, and for crowd-aware planning in \prettyref{sec:crowd_planner}. We call them \textit{CrowdMapper}, \textit{CrowdLocalizer} and \textit{CrowdPlanner}, respectively.

\subsection{CrowdMapper}
\label{sec:crowd_mapper}

In \textit{CrowdMapper}, the robot leverages the local crowd distribution around itself as the sensory measurement for mapping, which can be reliably collected nowadays using pedestrian detection and tracking~\cite{bertoni2019monoloco} technique, thanks to the progress of computer vision. In particular, at time step $t$, the robot's local observation about the crowd is $\{\mathbf{m}_{t-\Delta t:t}^i\}_{i=1}^N$, where $N$ is the number of pedestrians within the robot's perception range and $i$ denotes the pedestrian ID in the tracking system. 
$\mathbf{m}_{t-\Delta t:t}^i = [\bar{p}_x,  \bar{p}_y, l, \theta] \in \mathbb{R}^4$ is a vector encoding the $i$-th pedestrian's state in terms of position and velocity at time step $t$, where $l$ is the moving distance in $\Delta t$ duration, $\theta$ is the moving direction, and $\bar{p}_x,  \bar{p}_y$ are the starting coordinates for the movement. In this paper, we choose $\Delta t=\SI{1}{s}$ to make a good trade-off between tracking fast movements and capturing stable movement patterns in the crowd.

Given local observations $\{\mathbf{m}_{t-\Delta t:t}^i \}$, the robot next uses $K$-means clustering~\cite{KMEANS} to recover time-invariant crowd-flow patterns locally around. In particular, we first discretize the 2D plane into grid cells $\mathcal{C}_r$ at a certain resolution. Then, we assign $\mathbf{m}_{t-\Delta t:t}^i$ to different cells using the location $ [\bar{p}_x,  \bar{p}_y]$ as index. After that, we apply $K$-means to all velocities $[l,\theta]$ assigned to the same cell to identify the possible navigation directions in each cell. Since the number of clusters is unknown beforehand, inspired by the elbow method~\cite{KMEANS_Elbow}, we attempt the cluster number from $1$ to the maximum cluster number $N_c$, and choose the final cluster number as the one that does not reduce the total within-cluster sum of square (WSS) value dramatically or whose WSS is below a certain threshold $\text{WSS}_{m}$. Finally the robot can recover the local crowd-flow map. 

To construct a global crowd-flow map, the robot needs to fuse the local maps obtained in an online manner when moving around in the scenario. The entire \textit{CrowdMapper} pipeline is shown in \prettyref{fig:mapping_demo}. Our constructed crowd-flow map can be applied in many navigation tasks, including but not limited to the localization and social-aware planning tasks that will be described below. 

\subsection{CrowdLocalizer}
\label{sec:crowd_localizer}

\begin{figure}
\centering
\includegraphics[width=.5\linewidth]{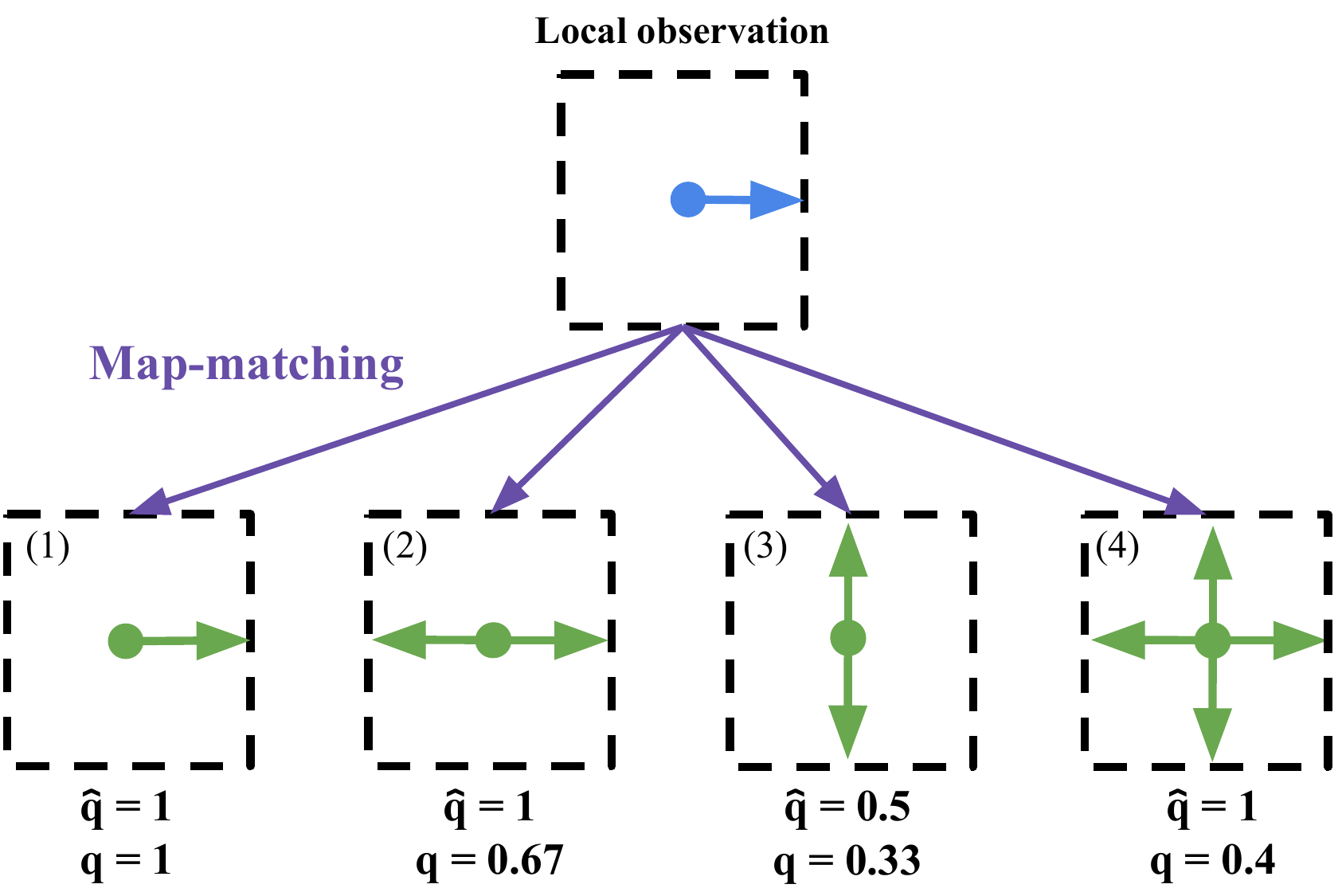}
\caption{Example of the flow-matching metric. Suppose that the observed crowd flow in the current grid cell is moving toward right, and there are four hypothesis cells in the map as shown in the bottom. The computed matching quality $q$ when the hyper-parameter $\gamma=0.5$ is listed for each candidate, where a larger $q$ indicates a higher matching quality.}
\label{fig:matching_example}
\end{figure}

We now demonstrate the robot can localize itself using the online constructed crowd-flow map instead of the traditional geometry map. Our approach is general and can be integrated in any traditional localization framework, but here we use the particle-filter based localization framework~\cite{PF_Localization} as an example to describe our method. 

In concrete, the particle-filter localization includes three repetitive steps: 1) generating $n_p$ hypothesis samples for robot locations, 2) computing the weight for each sample by evaluating the hypothesis quality in terms of the matching quality between the observation and map locally around the hypothesis, and 3) important sampling to update the hypothesis samples. When using the crowd-flow map for localization, we leave step 1) and 3) unchanged, because they are irrelevant to crowd-flow details. We modify step 2) by designing an appropriate metric to evaluate the matching quality between the local crowd observation and the crowd-flow map locally around a given location hypothesis. In particular, we first compute the minimum difference $\theta_m$ between an observed direction $\theta$ and all the possible directions in the grid cell where the location hypothesis locates. Next, we normalize $\theta_m$ to $[0,1]$ and get the score $\hat{q} = \frac{\pi - \theta_m}{\pi}$. In addition, among different hypothesis cells with the same $\hat{q}$ score, we prefer the cell with the minimum number of possible directions, since the corresponding matching is most robust. Thus, we discount $\hat{q}$ with the factor $\gamma \in [0, 1]$  and get the final matching score $q = \frac{1}{1 + (n - 1)\gamma} \hat{q}$,
where $n$ is the number of all directions in a hypothesis cell. Note that tuning $\gamma$ can trade off the preference between the diversity and error of the directions. We illustrate the flow-matching metric using a toy example in \prettyref{fig:matching_example}. Moreover, it is worth noting that our flow-matching metric is complementary with the traditional geometry-matching metric, and both metrics can be employed simultaneously to fuse the crowd and geometry information. In particular, the crowd-matching metric is desirable in highly dynamic scenarios like human crowds while the geometry-matching metric is more helpful in static scenarios.

\subsection{CrowdPlanner}
\label{sec:crowd_planner}

We now describe how to utilize rich social navigation information encoded in the constructed crowd-flow map to accomplish socially-compliant robotic navigation in crowds, which is non-trivial when using traditional geometry map. Although there exist some approaches that can take into account crowd flow into the navigation policy~\cite{aroor2019online, henry2010learning}, most of them assume an ideal localization, which is difficult to hold in real-world dense pedestrian scenarios. Our solution separates planning into two parts: in the long time horizon, the robot follows the global patterns in the existing crowd-flow map and plans a minimum cost trajectory connecting the robot's current location and its goal; in the short time horizon, the robot uses efficient re-planning to update the long horizon plan with online observed local crowd flow data, which is also beneficial for imperfect localization. Our planning algorithm is formulated as a search over the crowd-flow grid map, whose topology is defined by eight-neighborhood, i.e., the robot can move in eight different directions. Since imperfect localization needs frequent replanning, we choose D*~\cite{D_star_Lite} as the default planner, though other solvers such as A* can also be used. 

\begin{algorithm}
\caption{Navigation cost computation for a robot's possible movement}
\label{alg:cost_computation}
\begin{algorithmic}[1]
\State \textbf{Input:} The crowd-flow map $\mathcal{M}$, the robot's current cell location  $\mathbf{p}$, the robot's one possible movement $\mathbf{v}^{robot}$, weights $w_{rc}$ and $w_{lc}$ for resistance and lubricating costs respectively.

\State $Cost = \| \mathbf{v}^{robot} \|$
\State $[RC_{\mathbf{p}}, LC_{\mathbf{p}}] = [0, 0]$
\State // \textit{Compute the crowd-flow cost at the current location}
\For{$\mathbf{v}^{{flow}} \textbf{ in } \mathcal{M}(\mathbf{p})$}
\State $RC_{\mathbf{p}} \mathrel{+}= \mathbf{v}^{robot} \cdot (-\mathbf{v}^{{flow}})$, $LC_{\mathbf{p}} \mathrel{+}= \| \mathbf{v}^{robot} \times \mathbf{v}^{{flow}} \|$
\EndFor
\State $RC = RC_{\mathbf{p}}$, $LC = LC_{\mathbf{p}}$
\State // \textit{Sum up all cost with weights}
\State $Cost \mathrel{+}= w_{rc} \cdot RC + w_{lc} \cdot LC$
\State \textbf{Output:} $Cost$
\end{algorithmic}
\end{algorithm}

For the search-based planner, we need to specify cost and cost-to-go functions over the grid map for designing search heuristics. The cost function defines the transition cost when the robot moves between cells after executing a selected action, and the cost-to-go function estimates the shortest distance from the robot's current position to the goal. 
We use the same cost-to-go function as the geometry grid map, but we design a new cost function to encourage the robot following crowd flow patterns. In particular, besides the traditional cost penalizing the long-distance movement, we consider another two costs, the \textit{resistance cost} ($RC$) and \textit{lubricating cost} ($LC$). Intuitively, $RC$ penalizes the motions that are not consistent with the crowd flow, while $LC$ formulates the fact that moving at directions perpendicular to the possible crowd-flow directions in a cell is likely to meet obstacles and thus should be avoided. To compute the $RC$ cost for the robot at the position $\mathbf{p}$ taking the motion $\mathbf{v}^{robot}$, we project $\mathbf{v}^{robot}$ in the negative direction of every crowd-flow $-\mathbf{v}_{\mathbf{p}}^{flow_i}$ in the cell locating at $\mathbf{p}$. In this way, we get a scalar cost for each direction, i.e., $RC_{\mathbf{p}}^i = \mathbf{v}^{robot} \cdot (-\mathbf{v}_{\mathbf{p}}^{{flow}_i})$. We define the total $RC$ cost at $\mathbf p$ as $RC_{\mathbf{p}}=\sum_i RC_{\mathbf{p}}^i$. For $LC$ cost, we compute a vector $LC_{\mathbf{p}}^i$ as the cross product of $\mathbf{v}^{robot}$ and $\mathbf{v}_{\mathbf{p}}^{{flow}_i}$, i.e. $LC_{\mathbf{p}}^i = \| \mathbf{v}^{robot} \times \mathbf{v}_{\mathbf{p}}^{{flow}_i} \|$. The total $LC$ cost is defined as $LC_{\mathbf{p}} = \sum_i LC_{\mathbf{p}}^i$. 

Intuitively, the negative value of $RC$ implies that the robot movement is consistent with the crowd flow and thus the resulting navigation behavior is social-friendly, and vice-versa. $LC$ is always positive, and a higher $LC$ value means that the movement in this cell has lower lubrication, i.e. has a higher chance to collide with static obstacles. We finally combine these two costs with the traditional distance-based cost with suitable blending weights (i.e. $w_{rc}$ and $w_{lc}$) and the entire procedure for computing the crowd-aware navigation cost is shown in  \prettyref{alg:cost_computation}.

\section{Experiments}
\label{sec:exp}

In this section, we first describe the implementation details of our approach. Next, we analyze our mapping algorithm in different scenarios and then validate the applicability of the crow-flow map for localization and planning in a crowded scenario. Finally, we compare our proposed approach with two baseline methods. 

\subsection{Implementation Details}
Our simulation experiments are implemented on a PC with an Intel i7-9700 CPU and an NVIDIA RTX 2060 GPU. As shown in \prettyref{fig:mapping_demo}, Gazebo~\cite{Gazebo_simulator} is used for simulating the mobile Turtlebot platform with a \SI{5}{m} perception range; Menge~\cite{Menge} is used for simulating realistic crowd behaviors. And we summarize the hyper-parameters used in our experiments as follows. For \textit{CrowdMapper}, we set the elbow clustering threshold $\text{WSS}_m$ as $0.5$. For \textit{CrowdLocalizer}, we set the particle number $n_p$ to $500$ and the crowd-flow matching metric parameter $\gamma$ as $0.5$. For \textit{CrowdPlanner}, we set $w_{rc}$ to $1.0$ and $w_{lc}$ to $0.5$ because $RC$ should be more important than $LC$ for planning in crowds.

\begin{figure}
\centering
\begin{subfigure}{0.14\textwidth}
\centering
\includegraphics[width=1.0\linewidth]{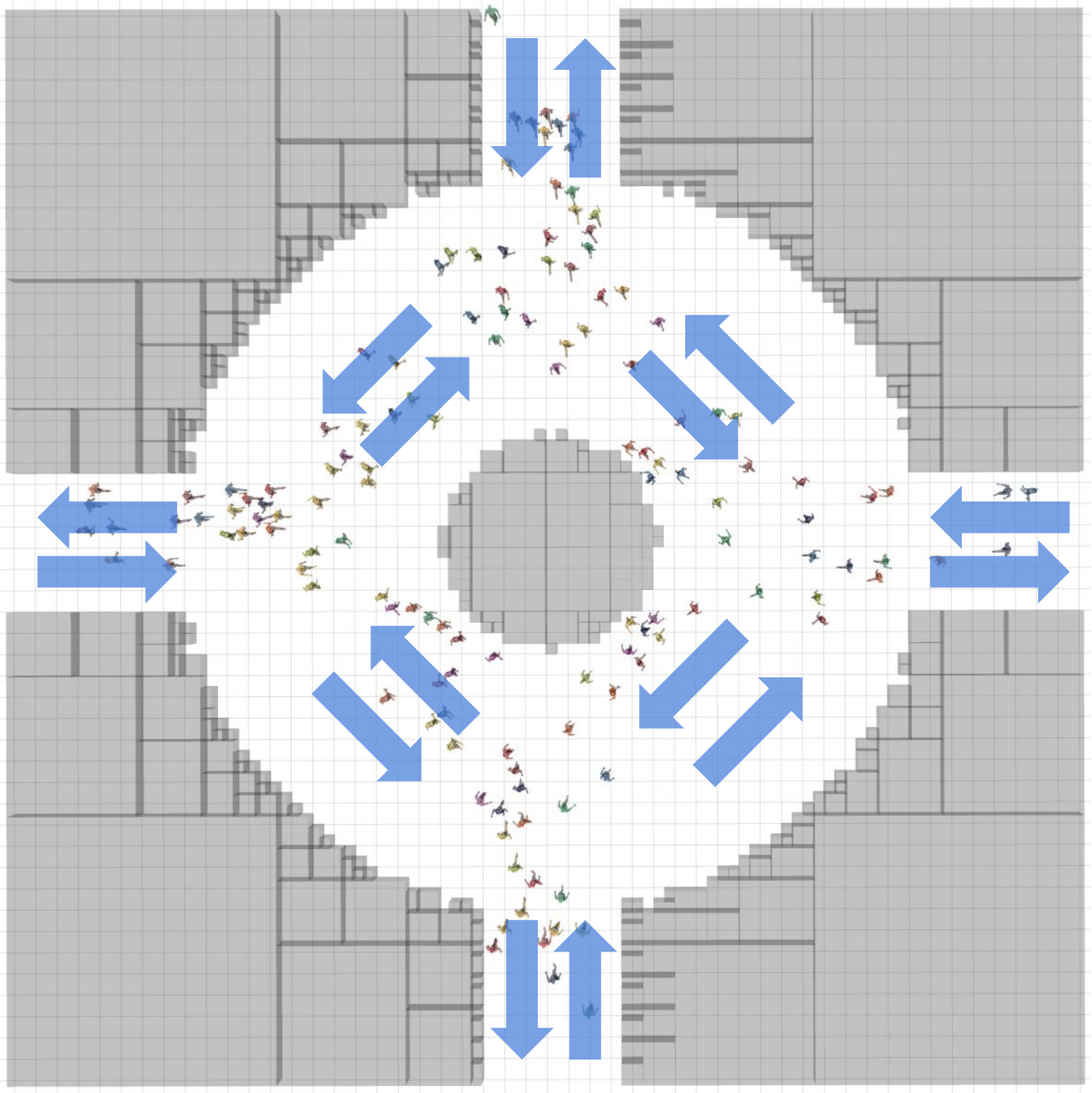}
\label{fig:fountain_scene}
\vspace{-0.5cm}
\caption{Fountain}
\end{subfigure} 
\begin{subfigure}{0.315\textwidth}
\raggedleft
\includegraphics[trim=0 30 0 30, clip, width=1.\linewidth]{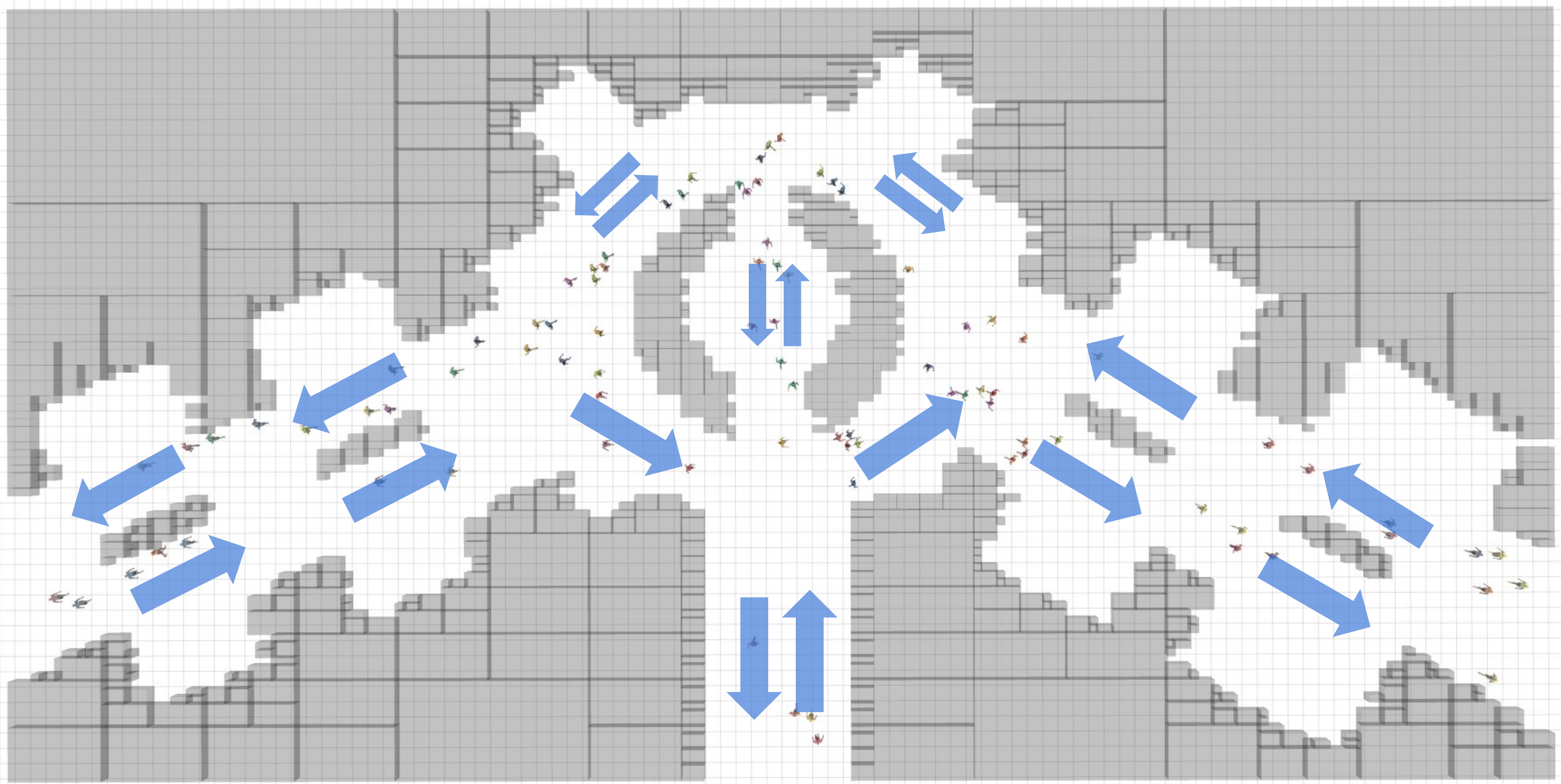}
\label{fig:conference_scene}
\vspace{-0.5cm}
\caption{Conference Venue}
\end{subfigure} 
\begin{subfigure}{0.52\textwidth}
\centering
\includegraphics[width=1.0\linewidth]{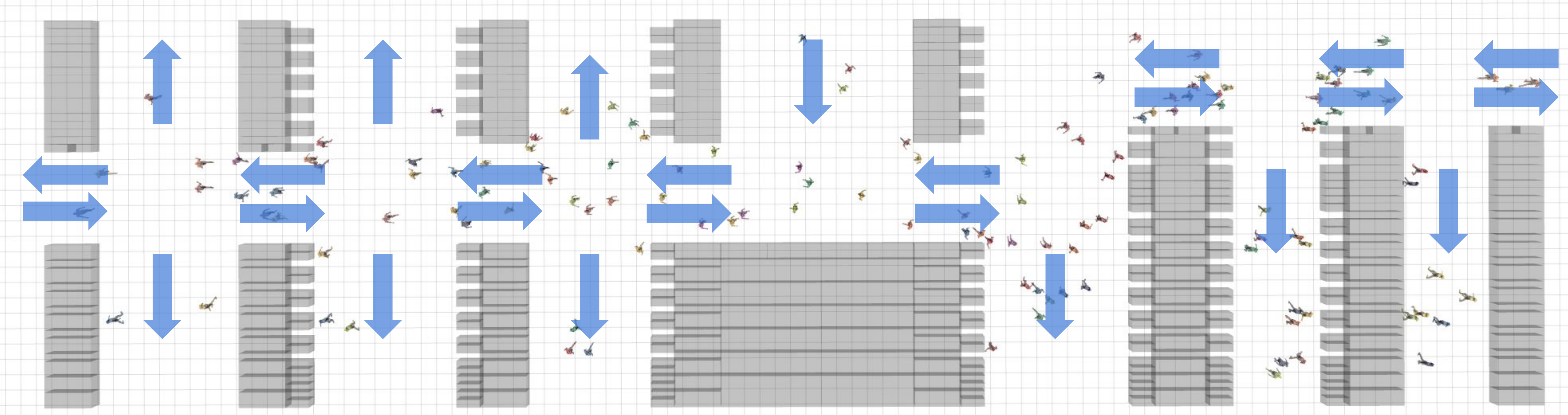}
\vspace{-0.5cm}
\label{fig:canteen_scene}
\caption{Canteen}
\end{subfigure}
\caption{Different benchmark scenarios. The blue arrows indicate crowd motion patterns. (a) is a $50 \times 50$ \SI{}{m^2} fountain scene whose crowd density is \SI{0.130}{/m^2}; (b) is a $100 \times 50$ \SI{}{m^2} conference venue scene whose crowd density is \SI{0.045}{/m^2}; (c) is a $100 \times 25$ \SI{}{m^2} canteen scene whose crowd density is \SI{0.081}{/m^2}.}
\label{fig:different_scenes}
\end{figure}

\subsection{Results for Crowd-Driven Mapping}
To evaluate the performance of our proposed algorithms, we set up several indoor and outdoor scenarios with different crowd densities, as shown in \prettyref{fig:different_scenes}. To quantitatively measure \textit{CrowdMapper}'s quality, we use the crowd information captured by an imaginary bird-view camera that globally and perfectly monitors the entire scenario to compute the ground-truth crowd flow field as the baseline. The errors between the baseline and our reconstructed crowd-flow map are used to investigate whether the time-invariant movement patterns of pedestrians are well captured. For the error metric, since we are measuring the difference in flow, we borrow the flow-matching metric in \prettyref{sec:crowd_localizer}. However, since the direction variance in a cell is not important, we use $\hat{q}$ rather than $q$ as the error metric for a single grid cell. We finally calculate the average error over all cells as the error over the entire grid-based map. 

\begin{figure}
\centering
\begin{subfigure}{0.39\textwidth}
\centering
\includegraphics[width=1.0\linewidth]{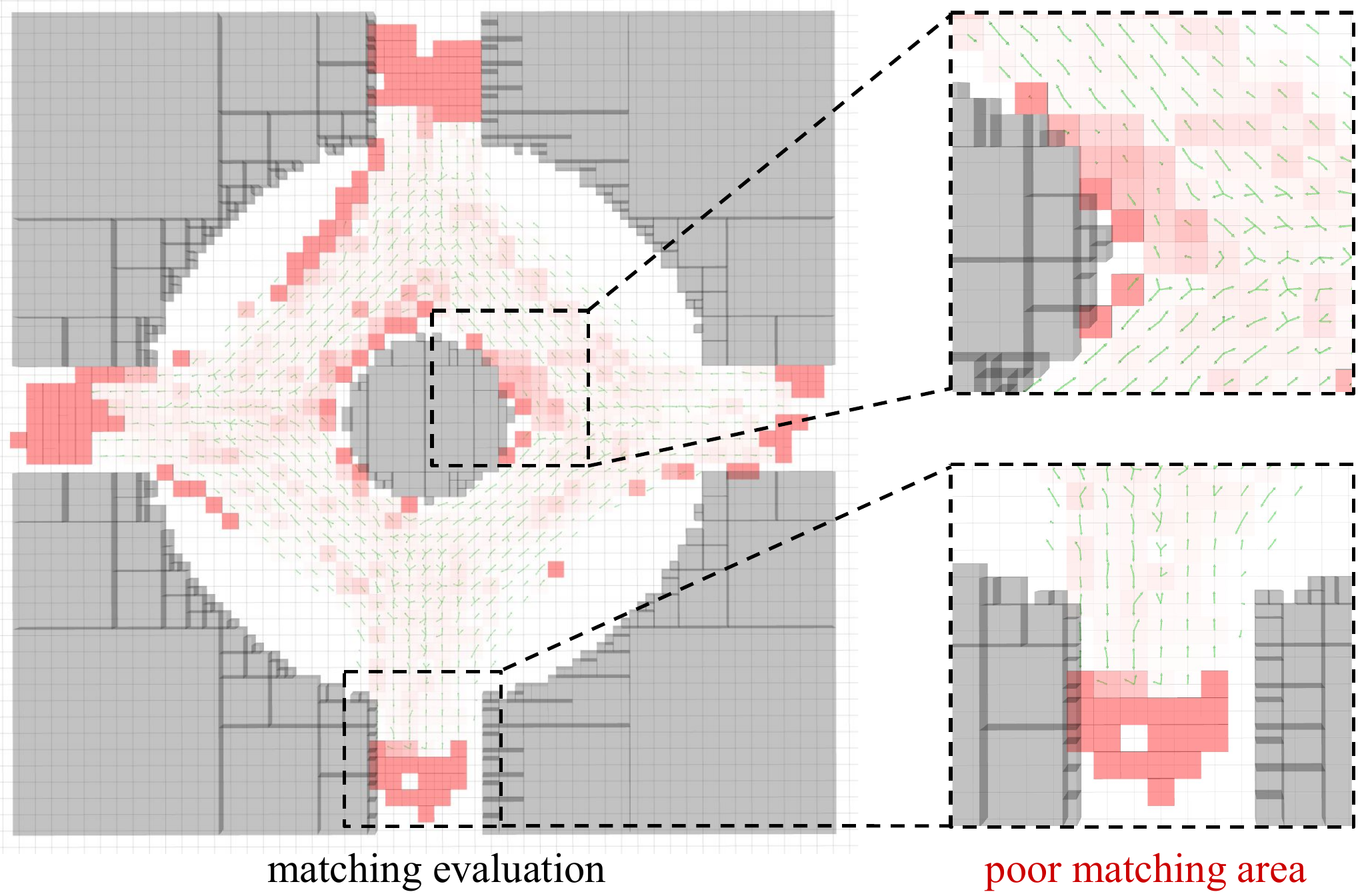}
\caption{ }
\label{fig:poor_matching_demo}
\end{subfigure}
\begin{subfigure}{0.6\textwidth}
\centering
\includegraphics[width=0.85\linewidth]{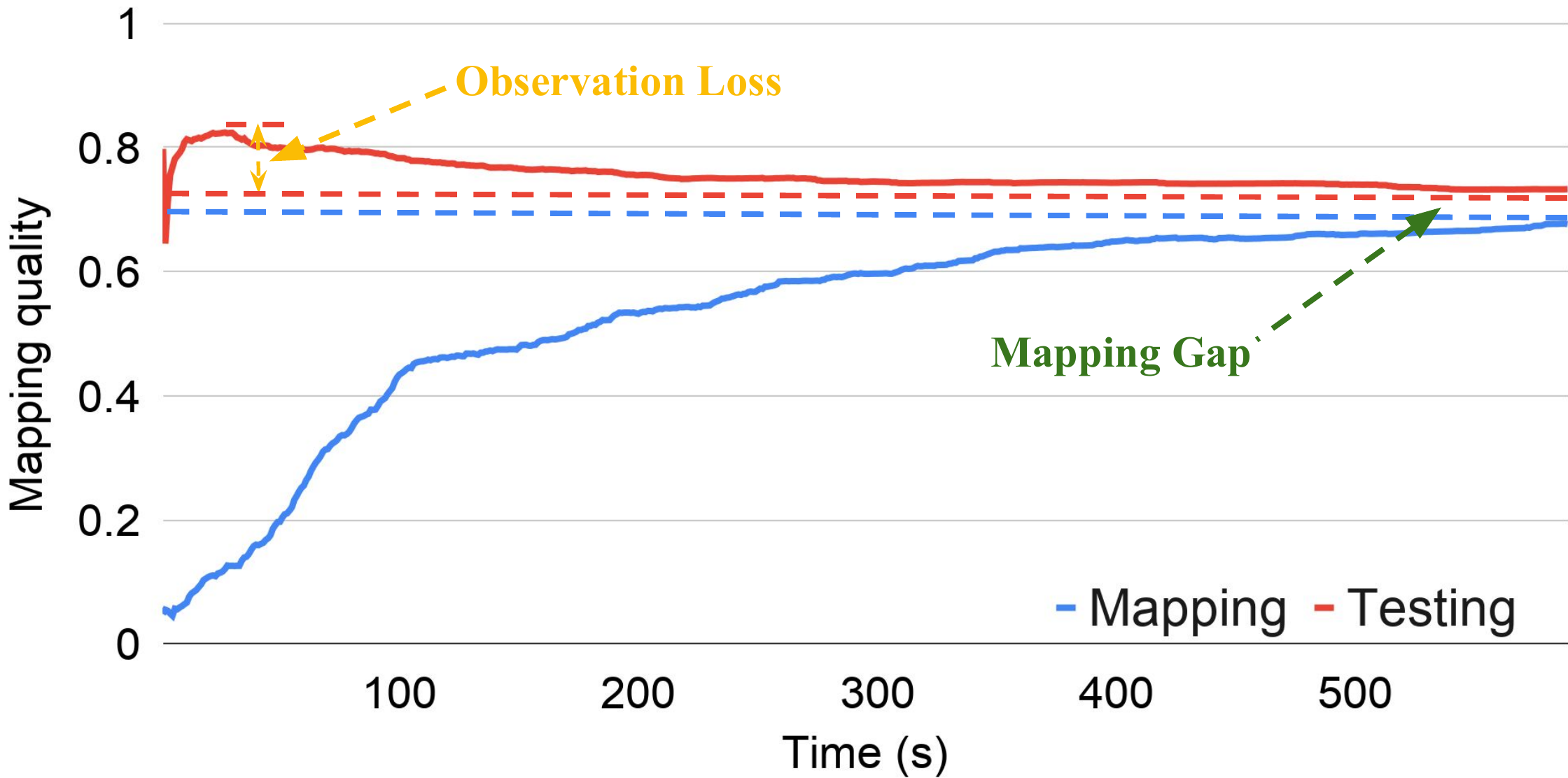}
\caption{ }
\label{fig:mapping_vs_testing}
\end{subfigure}
\caption{(a) Heatmap for fountain scene's map quality. The darkness of the red color indicates the loss magnitude. (b) The map quality of the mapping and testing stages over different time steps.}
\end{figure}

We analyze the mapping result in two separate stages, the mapping stage and testing stage, as shown in \prettyref{fig:mapping_vs_testing}. In the mapping stage, the mapping quality increases when the robot explores the scenarios but eventually it converges to a fixed value, marked by the blue dash line in \prettyref{fig:mapping_vs_testing}. Apparently the mapping quality is not perfect (i.e., $< 1$), which can be explained by two main challenges. First, we only discretize the crowd flow in a cell into eight directions and thus the robot cannot accurately model regions with complex crowd flow. Second, the robot only has partial observation to the scenario and may miss areas with a few pedestrians or crowds appearing in a period when it is absent. 

After the mapping stage completes, we compare the obtained global crowd-flow map with the pedestrian motion field real-time captured by the imaginary bird-view camera for a series of frames. As shown in \prettyref{fig:mapping_vs_testing}, the mapping quality is unstable at the beginning, and after a while gradually decreases until it converges to another fixed value marked by the red dashed line in \prettyref{fig:mapping_vs_testing}. The slow declined tail in the procedure can also be explained by robot's partial observation, since the global camera can observe crowd flow appearing in regions with rare human activity while the robot cannot. We provide more detailed analysis about this phenomenon in the next paragraph. Note that there is a small gap between the red and blue dashed lines. This is because the robot will affect the movement of surrounding pedestrians during the mapping stage but not during the testing stage.

Now we provide a detailed analysis about the error distribution using the fountain scenario as an example. We plot in \prettyref{fig:poor_matching_demo} the heatmap of the converged quality loss (i.e. $1$ - matching quality) in the testing stage. We can observe two main types of poor matching regions. First, in some areas, the robot can hardly observe any crowd, which leads to a significant inconsistency with the observation from the bird-view camera. These areas are often near the boundary of crowd-flow map, as shown by the zoomed-in region in the bottom-right of \prettyref{fig:poor_matching_demo}. Second, the robot indeed observes the crowd motion in some areas, but the observed data is insufficient for computing accurate time-invariant features, e.g., the region shown in the zoomed top-right in \prettyref{fig:poor_matching_demo}. However, as shown in the heatmap, our mapping approach can obtain a high quality crowd-flow map correctly reflecting the layout of the entire scenario.

\begin{figure}
\centering
\begin{subfigure}{0.48\textwidth}
\centering
\includegraphics[trim=80 120 30 0, clip, width=1.0\linewidth]{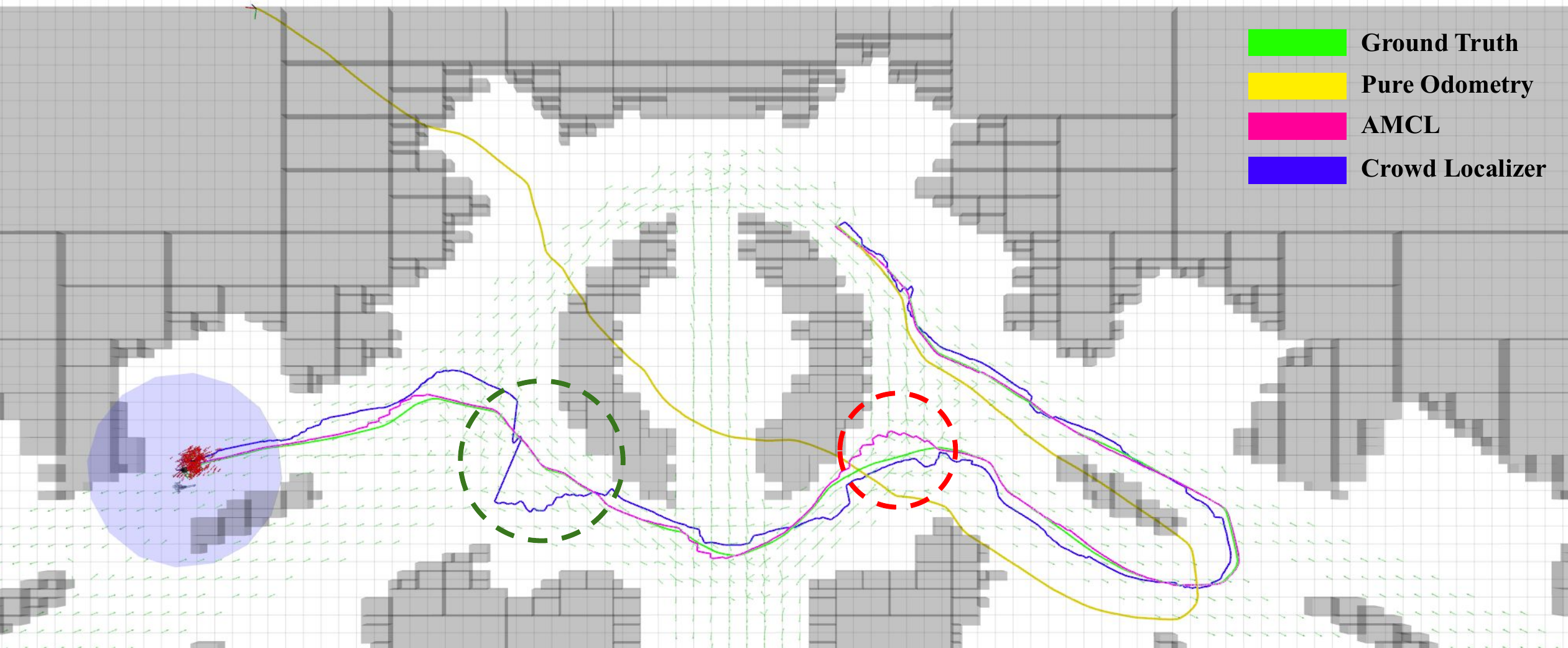}
\caption{Trajectories generated by various methods}
\label{fig:traj_conference}
\end{subfigure}
\begin{subfigure}{0.48\textwidth}
\centering
\includegraphics[width=1.0\linewidth]{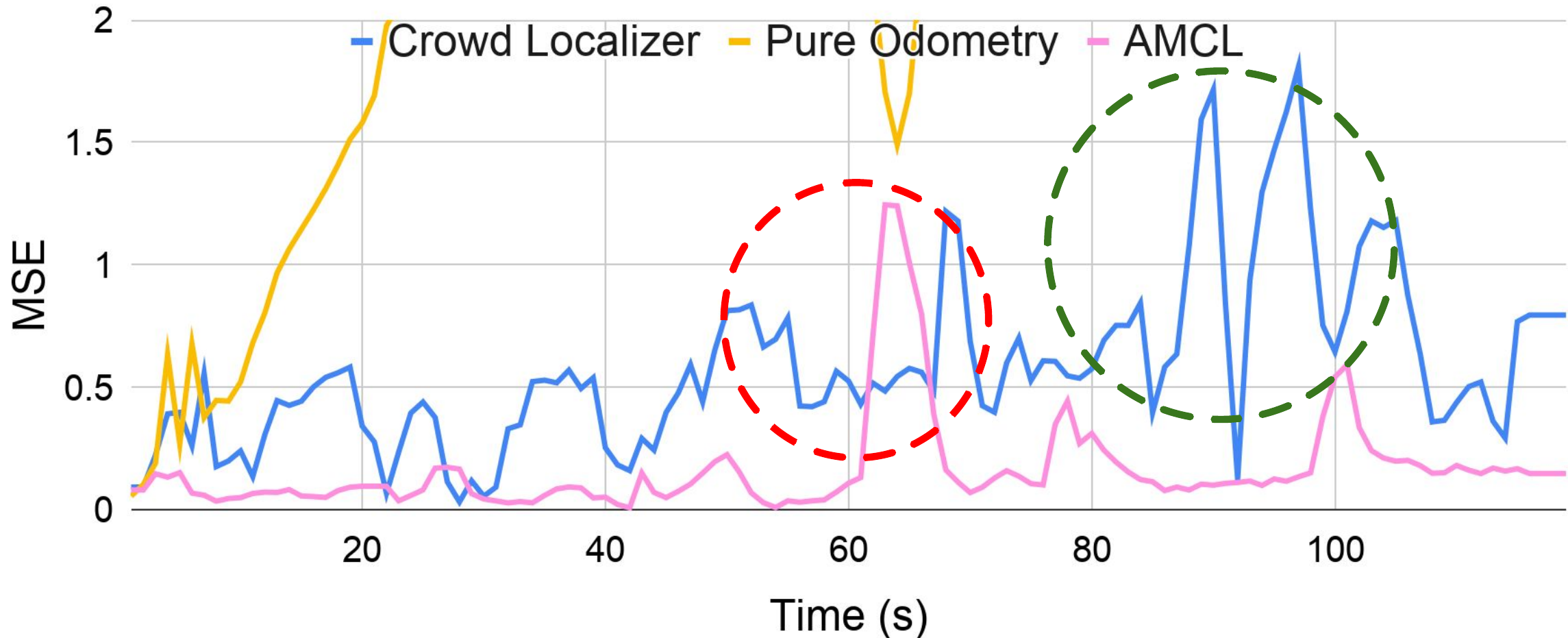}
\caption{Localization error}
\label{fig:localization_curve}
\end{subfigure}
\caption{A localization example in the \textit{Conference Venue} scene. }
\label{fig:localization_example}
\end{figure}

\subsection{Results for Crowd-Driven Localization and Planning}
\label{sec:localization_planning}

After constructing a crowd-flow map, we next investigate whether \textit{CrowdLocalizer} and \textit{CrowdPlanner} can leverage the map to achieve localization and planning. Here we evaluate these two modules separately, i.e., evaluating first the localization accuracy, and then the planning performance assuming perfect localization. We will discuss planning under imperfect localization in \prettyref{sec:complete}.  

We compute the localization accuracy as the Mean Square Error (MSE) between the ground truth and estimated paths. To better understand the effectiveness of \textit{CrowdLocalizer}, we compare it with two baseline methods: the pure odometry method and a geometry-based particle filter method AMCL~\cite{ACML-paper}, both are implemented in ROS. 

\begin{figure}
\centering
\includegraphics[width=1.0\linewidth]{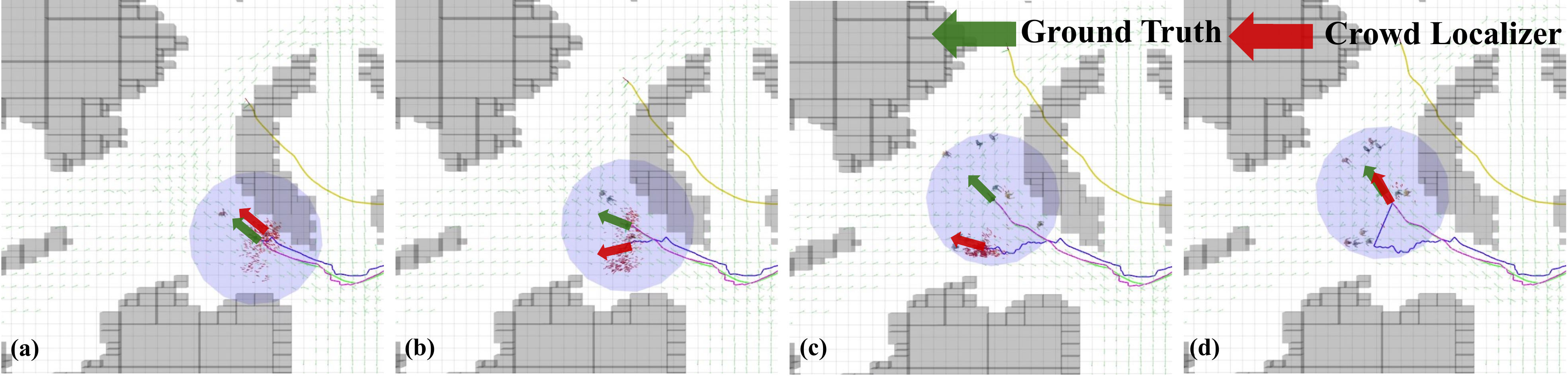}
\caption{Our localization quickly adjusts the pose estimation when crowds are observed.} 
\label{fig:crowd_localizer_correct_error}
\end{figure}

One qualitative comparison between these methods is shown in \prettyref{fig:localization_example} on the conference venue benchmark. We can observe that both \textit{CrowdLocalizer} and AMCL can well amend odometry sensor's cumulative errors. However, since they use different sensory observations (geometry for AMCL and crowd for \textit{CrowdLocalizer}), their localization error distributions are different. For AMCL, the localization drift may occur around the intersection of crowds where people occlude most static landmarks, as shown by the red dashed circle in \prettyref{fig:localization_example}. In contrast, the localization drift of \textit{CrowdLocalizer} usually appears in areas with sparse pedestrians, as shown by the green dashed circle in \prettyref{fig:localization_example}. When the observed pedestrian number increases, the drift can be effectively corrected, as shown in \prettyref{fig:crowd_localizer_correct_error}. 

Similar results are also observed in quantitative experiments, where we let the robot move for $300$ seconds in the benchmark scenarios and compute the average MSE of each trajectory. The results in \prettyref{tab:pure_localization} indicate that, as the crowd density increases, the localization error of AMCL gradually increases while the error of \textit{CrowdLocalizer} is relatively stable and even slightly decreases. In general, traditional localization approaches is more advantageous in regions with a few people, while our method outperforms in dense crowds. By combining traditional approaches, our method can solve localization tasks in general scenarios with varying crowd density.

\begin{table}
\centering
    \caption{The MSE of trajectories in the testing scenarios: Fountain, Canteen, and Conference Venue.}
    \label{tab:pure_local}
    \fontsize{7.5}{7.5}\selectfont
    \footnotesize
    \bgroup
    \def\arraystretch{1.3}
\begin{tabular}{l|ccc}
\hline
Method & \multicolumn{1}{c}{\textit{Fountain}} & \multicolumn{1}{c}{\textit{Canteen}} & \multicolumn{1}{c}{\textit{Conf. Venue}} \\ \hline
Pure Odometry & 3.470 & 18.915 & 6.188 \\
AMCL & 9.029 & 0.610 & \textbf{0.105} \\
CrowdLocalizer & \textbf{0.586} & \textbf{0.599} & 0.613 \\ \hline
\end{tabular}%
\egroup
\label{tab:pure_localization}
\end{table}

\begin{figure}
\centering
\begin{subfigure}{0.163\textwidth}
\centering
\includegraphics[trim=30 70 0 70, clip, width=1.\linewidth]{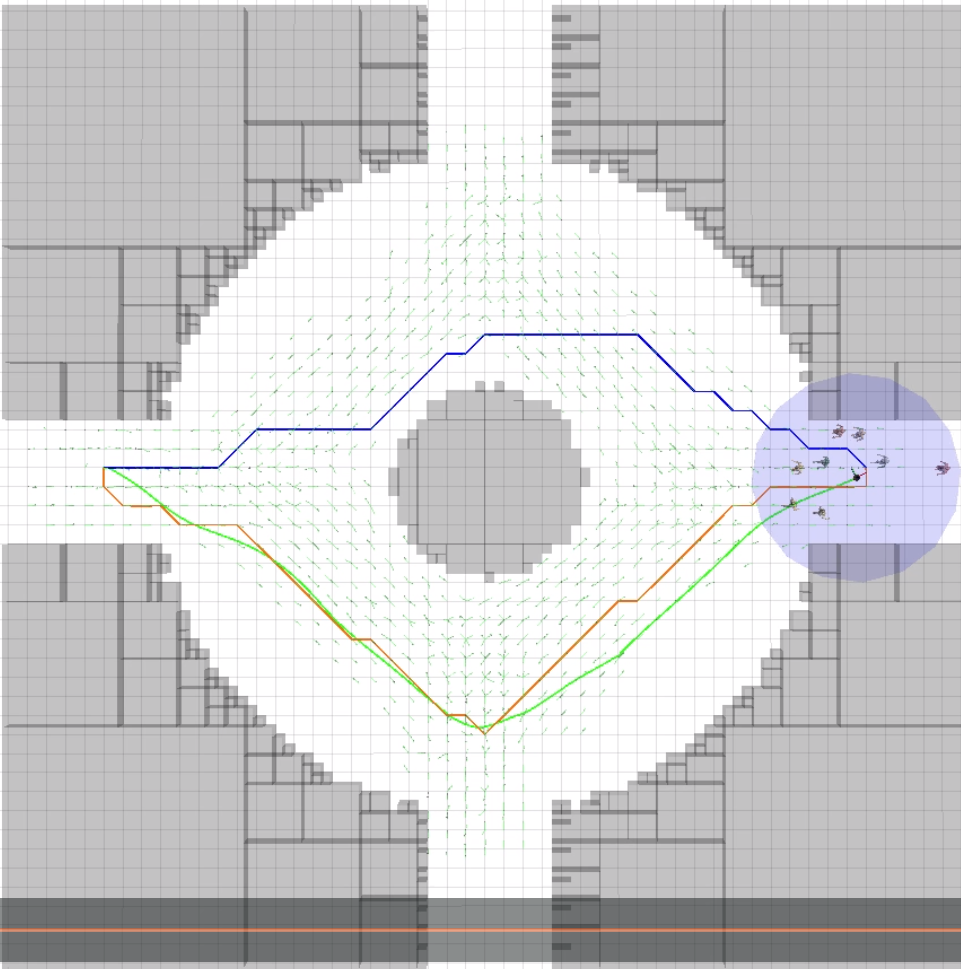}
\caption{Fountain}
\label{fig:fountain_traj}
\end{subfigure} 
\begin{subfigure}{0.317\textwidth}
\raggedleft
\includegraphics[trim=190 15 200 40, clip, width=1.\linewidth]{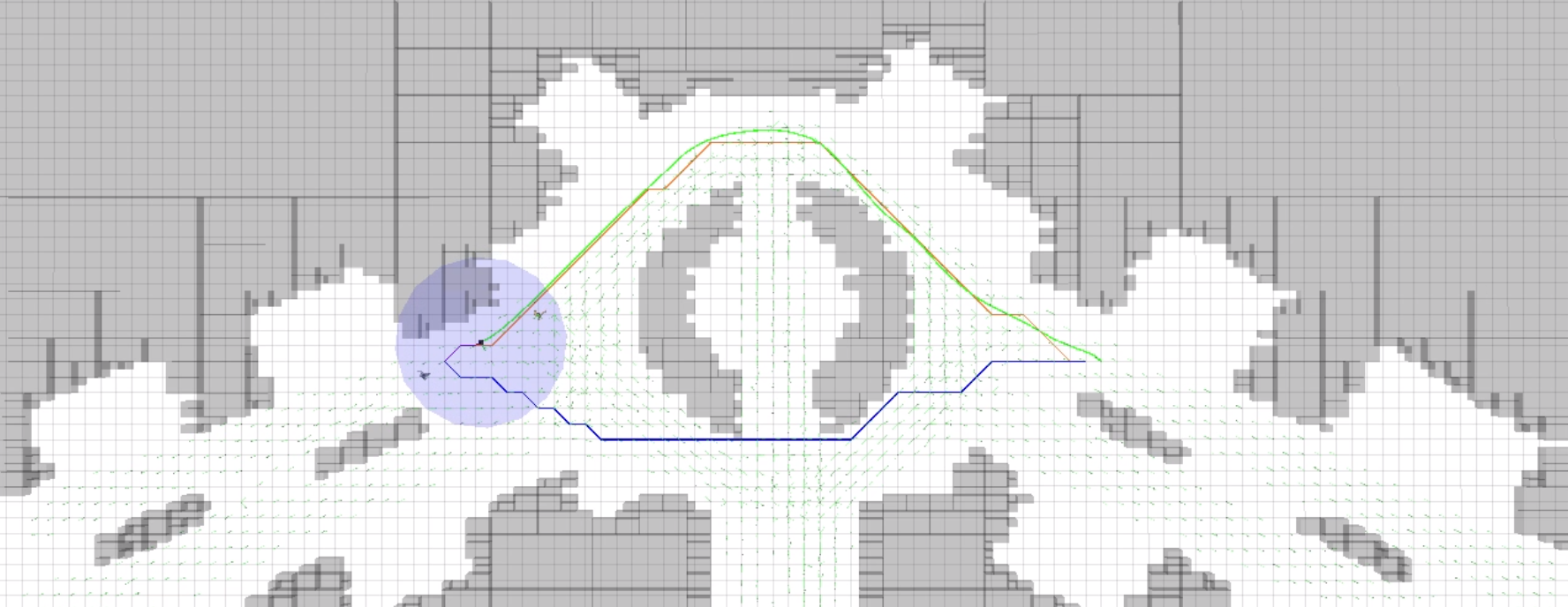}
\caption{Conference Venue}
\label{fig:conference_traj}
\end{subfigure} 
\begin{subfigure}{0.48\textwidth}
\centering
\includegraphics[trim=0 80 0 50, clip, width=1.0\linewidth]{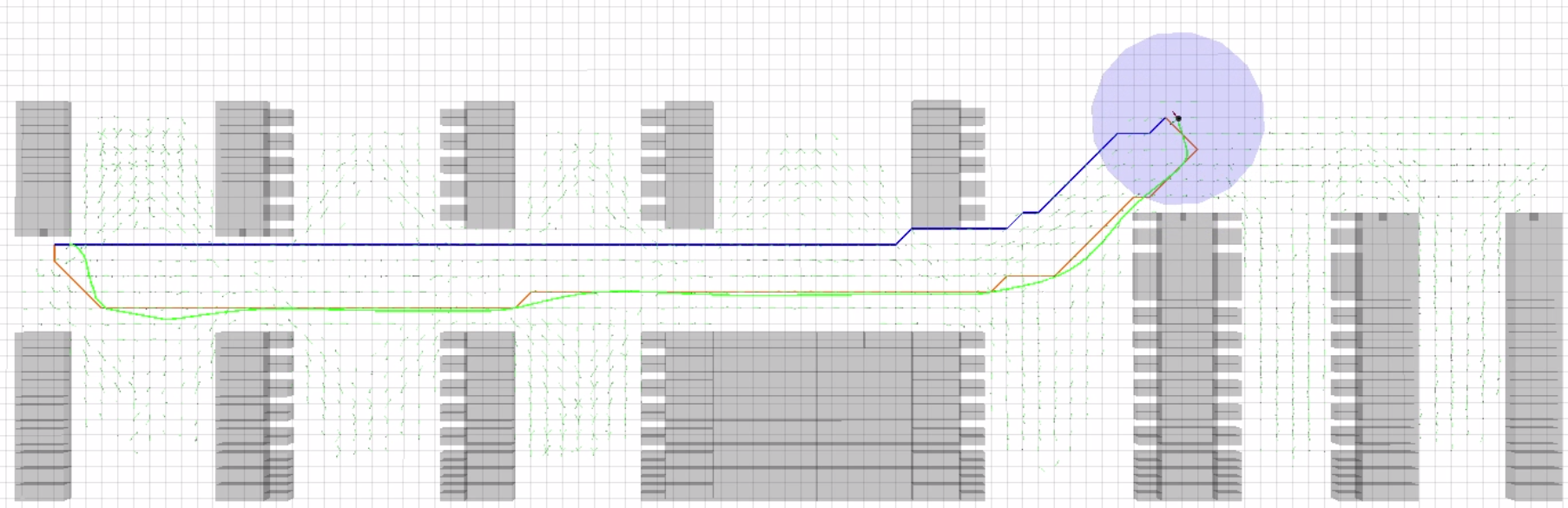}
\caption{Canteen}
\label{fig:canteen_traj}
\end{subfigure}
\caption{Trajectories planned by different approaches. The blue curve is the result of \textit{A* (shortest)}; the orange curve is result of \textit{A* (social)}, and is also the plan for \textit{CrowdPlanner} using D* at time $0$; the green curve is the final trajectory of \textit{CrowdPlanner} after D* replanning. }
\label{fig:different_scenes_traj}
\end{figure}

\begin{figure}
\centering
\includegraphics[trim={2.5cm 1cm 2.5cm 0.6cm},clip, width=0.8\linewidth]{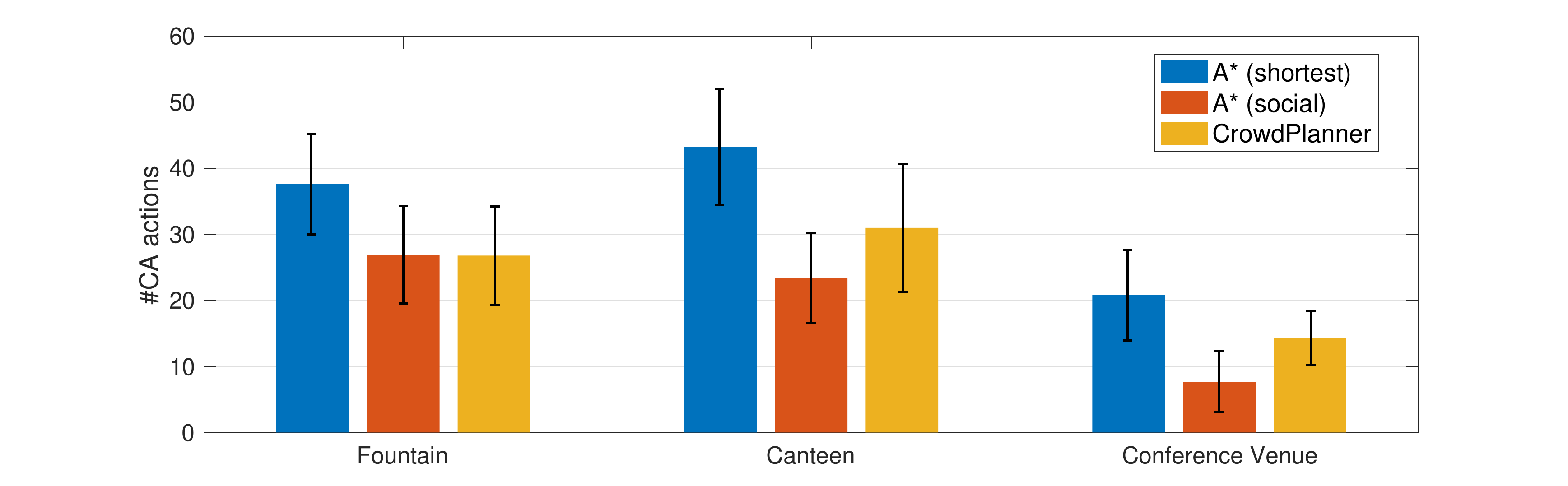}
\caption{The numbers of collision avoidance actions in trajectories planned using different approaches in benchmark scenarios (from right to right: fountain, conference, and canteen.)}
\label{fig:pure_planning_bar}
\end{figure}

We then evaluate the performance of the planning module in term of social-friendly crowd navigation under perfect localization. Given a planning result, the robot will follow the plan and meanwhile actively avoid collisions with nearby pedestrians using a state-of-the-art reactive collision avoidance policy~\cite{fan2019getting}. A high-quality plan is expected to minimize the inter-disturbance with the crowd. Thus we use the number of the robot's active collision avoidance actions (i.e. \#CA actions) during the navigation to quantify whether the planned path is socially compliant. Motivated by the fact that a plan passes through open space, which has the minimum inter-disturbance with the crowd, should be a smooth path with limited acceleration, we use the acceleration in the path over a short period of $t_{traj}$ as a signal to detect CA actions. If the acceleration exceeds a predefined threshold $\dot{v}_{thres}$, the robot is considered as taking a CA action at that time. Through our experiments, we find that CA actions can be well detected with $t_{traj} = \SI{0.5}{s}$ and $\dot{v}_{thres} = \SI{0.15}{m/s^2}$. Furthermore, we compare the shortest trajectory planned by A*~\cite{A_Star_Algorithm} (i.e. \textit{A* (shortest)}), with trajectories that take into account the crowd-flow map but using different solvers A* (i.e. \textit{A* (social)}) and D* (i.e. \textit{CrowdPlanner}).

\begin{table*}
\centering
\footnotesize
\caption{Comparison experiments measured by the success rate and the number of executed collision avoidance actions (mean/std).}
\resizebox{1.0\textwidth}{!}{
\begin{tabular}{c|l|c|r|c|c|cr}
\hline
\multicolumn{2}{c|}{Method} & \multicolumn{2}{c|}{\textit{Fountain}} & \multicolumn{2}{c|}{\textit{Canteen}} & \multicolumn{2}{c}{\textit{Conf. Venue}} \\ \hline
\multicolumn{1}{c|}{Localizer} & \multicolumn{1}{c|}{Planner} & \multicolumn{1}{c|}{Success rate} & \multicolumn{1}{c|}{\# CA actions} & \multicolumn{1}{c|}{Success rate} & \multicolumn{1}{c|}{\# CA actions}  & \multicolumn{1}{c|}{Success rate} & \multicolumn{1}{c}{\# CA actions}  \\ \hline
\multirow{3}{*}{AMCL} & A* (shortest) & 92\% & 39.26 (11.34) & \textbf{90\%} & 36.20 (7.83) & \multicolumn{1}{c|}{\textbf{100\%}} & 22.00 (8.34) \\
 & A* (social) & 34\% & \textbf{34.38  (8.11)} & 86\% & 31.91 (8.73) & \multicolumn{1}{c|}{\textbf{100\%}} & \textbf{11.40 (6.24)} \\
 & CrowdPlanner & 52\% & 43.87 (12.92) & 80\% & 40.36 (7.69) & \multicolumn{1}{c|}{\textbf{100\%}} & 19.16 (8.78) \\ \hline
\multirow{3}{*}{CrowdLocalizer} & A* (shortest) & 90\% & 54.62 (14.31) & 82\% & 40.96 (12.92) & \multicolumn{1}{c|}{90\%} & 25.06 (7.44) \\
 & A* (social) & 86\% & 44.71 (15.36) & 84\% & 30.71 (10.56) & \multicolumn{1}{c|}{92\%} & 11.85 (10.14) \\
 & CrowdPlanner & \textbf{96\%} & 40.38 (12.08) & \textbf{90\%} & \textbf{29.58 (9.64)} & \multicolumn{1}{c|}{84\%} & 25.44 (7.68) \\ \hline
\end{tabular}%
}
\label{tab:comparison_exp}
\end{table*}

Since the crowd motion is random in simulation, we run each planning algorithm for $50$ times in each testing scenario and then compare the different trajectory plans, as demonstrated in \prettyref{fig:different_scenes_traj}. We can observe that the planners (i.e. \textit{CrowdPlanner} and \textit{A* (social)}) considering the social flow is prone to generate a crowd-following trajectory, though their paths can be longer than that of \textit{A* (shortest)} planner that is not social aware. In addition, we count \#CA actions a robot will execute during the navigation and the results are summarized in \prettyref{fig:pure_planning_bar}. We can observe that the social-aware planners outperform the \textit{A* (shortest)} planner.

\subsection{Comparison of Complete Navigation Systems}
\label{sec:complete}
Finally, we compare our crowd-based navigation framework with two baseline navigators to evaluate the overall performance. Similar to our previous experiments, we employ the \#CA actions to measure the social compliance of a navigation behavior. In addition, we also compute the navigation success rate in terms of arriving goal without collisions to evaluate the navigator's robustness. 

From the results in \prettyref{tab:comparison_exp}, we find that the AMCL localizer is sensitive to crowd density and planner type, whose navigation success rate is significantly downgraded in dense crowd scenarios such as Fountain and Canteen. The only exception is when AMCL is working with \textit{A* (shortest)} whose navigation behavior (as shown by the blue trajectory in \prettyref{fig:fountain_traj}) appears to be similar to a coastal planner \cite{Coastal_Planner}, i.e., whenever the robot is about to lose its localization, it can walk towards nearby static landmarks to correct the error. In contrast, the success rate of \textit{CrowdLocalizer} is stable in all three scenes, though in Conference scene with sparse pedestrians, it performs less perfect than AMCL. Besides, \textit{CrowdPlanner} can dynamically follow the movement of nearby pedestrians and thus can improve the success rate compared to the \textit{A* (social)} planner. In general, the combination of \textit{CrowdLocalizer} and \textit{CrowdPlanner} is superior in scenes with dense crowds, while in less crowded scenes, AMCL with social-aware planner achieves better results.

\subsection{Real-robot Experiments in a rush hour crowd}
\label{sec:real_exp}
We have deployed our crowd-driven navigation system on a wheeled robot to validate its feasibility and performance in the real world. The experiments are conducted on a university campus at rush hour to make a challenging and real scenario with dense pedestrians. We first teleoperate the robot to navigate within the crowd to construct the crowd-flow map. Next, we test the performance of our proposed localization and planning methods. To make the navigation more robust, we also complement our crowd-driven methods with the traditional SLAM method~\cite{legoloam2018shan}, as we discussed in~\ref{sec:localization_planning}. For more details, please refer to our supplementary multimedia materials.

\section{Conclusion}

We propose to utilize the movement information of pedestrians, which is commonly treated as harmful noises in traditional SLAM, for mapping, localization, and social-aware planning. Compared with the geometry map in traditional SLAM, our crowd-flow map provides a new perspective for robots better understanding its surrounding dynamic scenarios. Our approach has demonstrated preliminary but promising navigation performance on both simulated and real-world scenarios.

\footnotesize{
\bibliographystyle{plainnat}
\bibliography{references}
}
\end{document}